\documentclass[conference]{IEEEtran}

\IEEEoverridecommandlockouts
\usepackage{cite}
\usepackage{amsmath,amssymb,amsfonts}
\usepackage{algorithmic}
\usepackage{graphicx}
\usepackage{textcomp}
\usepackage{subcaption}
\usepackage[table,xcdraw]{xcolor}
\usepackage{tabularx}
\usepackage{multirow}
\usepackage[hidelinks]{hyperref}
\usepackage{tikz}
\usepackage{aveasmargins}

\def\BibTeX{{\rm B\kern-.05em{\sc i\kern-.025em b}\kern-.08em
    T\kern-.1667em\lower.7ex\hbox{E}\kern-.125emX}}
\begin{document}

\bstctlcite{IEEEexample:BSTcontrol}
\setlength{\columnsep}{0.24in}

\aveasSetMargins{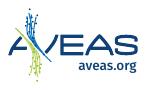}
\aveasSetIEEEFoot{2023}
\aveasSetIEEEHead{N. Neis and J. Beyerer, "Literature Review on Maneuver-Based Scenario Description for Automated Driving Simulations," 2023 IEEE Intelligent Vehicles Symposium (IV), Anchorage, AK, USA, 2023, pp. 1-8}{10.1109/IV55152.2023.10186545}

 \title{\bfseries{Literature Review on\\Maneuver-Based Scenario Description for\\Automated Driving Simulations}}

\author{Nicole Neis$^{1}$, Juergen Beyerer$^{2,3}$\vspace{-9pt}%

\thanks{*This publication was written in the context of the AVEAS research project (www.aveas.org), funded by the German Federal Ministry for Economic Affairs and Climate Action (BMWK) within the program ``New Vehicle and System Technologies''.}%
\thanks{
\raggedright
$^{1}$Porsche Engineering Group GmbH, 71287 Weissach, Germany,
{\tt\footnotesize nicole.neis@porsche-engineering.de}}%
\thanks{
\raggedright
$^{2}$Fraunhofer IOSB, 76131 Karlsruhe, Germany}%
\thanks{
\raggedright
$^{3}$Karlsruhe Institute of Technology (KIT), Vision and Fusion Laboratory (IES), 76131 Karlsruhe, Germany
}%
}

\maketitle

\begin{abstract}
The increasing complexity of automated driving functions and their growing operational design domains imply more demanding requirements on their validation. Classical methods such as field tests or formal analyses are not sufficient anymore and need to be complemented by simulations. For simulations, the standard approach is scenario-based testing, as opposed to distance-based testing primarily performed in field tests. Currently, the time evolution of specific scenarios is mainly described using trajectories, which limit or at least hamper generalizations towards variations. As an alternative, \emph{maneuver-based approaches} have been proposed. We shed light on the state of the art and available foundations for this new method through a literature review of early and recent works related to maneuver-based scenario description. It includes related modeling approaches originally developed for other applications. Current limitations and research gaps are identified.\end{abstract}


\section{Introduction}

\begin{figure*}[htbp]
\centering
\includegraphics[width=\textwidth]{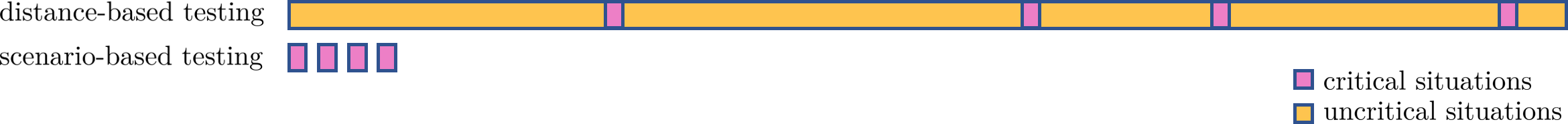}	
\caption{\centering Distance- vs. scenario-based testing: scenario-based testing can concentrate on critical situations while long distances of uncritical situations have to be passed before encountering the critical ones in case of distance-based testing}
\label{fig:testing}
\end{figure*}

\IEEEPARstart{R}{esearch} in automated driving (AD) functions aims at automated systems taking over an increasing share of the dynamic driving task with growing operational design domains and a decreasing need for the driver to take over control over the vehicle \cite{reschkaFertigkeitenUndFaehigkeitengraphen2017,VANBRUMMELEN2018384}. The fatal Tesla accident of 2016 \cite{plazaCollisionCarOperating2017} showcased the difficulties related to the validation of such AD functions under increasing complexity: the accident was the result of various factors coinciding in a singular scenario: the system was not designed to, and did not detect a crossing semi-trailer truck under the given circumstances. Furthermore, neither the Tesla driver nor the truck driver paid sufficient attention. In addition to that, the means by which the Tesla monitored the driver's attention and the system's reaction to it were found inappropriate. This highlights that testing solely based on the specification of the components is not sufficient \cite{sipplIdentificationRelevantTraffic2020,wilhelmFunktionaleSicherheitUnd2015}. Instead, the system as a whole needs to be tested in its realistic, diverse operational environment. Classically, field tests performed this task. They are mainly based on the principle of distance-based testing \cite{montanariManeuverbasedResimulationDriving2021, wachenfeldFreigabeAutonomenFahrens2015}: a certain amount of kilometers is driven under conditions in which the AD is expected to work in order to draw a statistical conclusion on a system's safety. However, to statistically prove that a highway pilot function operating on German highways is at least as safe as a human driver, $6.6\times 10^9$ km must be driven on German highways~\cite{wachenfeldFreigabeAutonomenFahrens2015}. Therefore, classical validation methods such as field tests \cite{bachMethodenUndAnsaetze2018,knappCodePracticeDesign2009} need to be complemented by simulations. A critical factor driving up the amount of kilometers that theoretically need to be driven is the low probability of situations that bring the system to its limits \cite{wachenfeldFreigabeAutonomenFahrens2015}. Simulations enable the step from distance- to scenario-based testing \cite{schuldtBeitragFuerMethodischen2016, elrofaiSCENARIOBASEDSAFETYVALIDATION2018, pfefferSzenariobasierteSimulationsgestuetzteFunktionale2020, putzSystemValidationHighly2017, riedmaierSurveyScenarioBasedSafety2020, sipplIdentificationRelevantTraffic2020}. As illustrated in Fig.~\ref{fig:testing}, scenario-based testing can explicitly focus on the critical situations. The long chain of uncritical situations that needs to be passed in case of distance-based testing can be left out. The key challenge with scenario-based testing is the selection of a scenario catalogue. The number and character of the scenarios to test is still an open question and the content of numerous research papers \cite{amersbachFunktionaleDekompositionBeitrag2019, zhangFindingCriticalScenarios2021,pegasus-projektpartnerSchlussberichtFuerGesamtprojekt2020}. However, as outlined in \cite{grundlFehlerUndFehlverhalten2005}, already the validation based on a limited scenario catalogue can increase the safety of an AD function, by reducing risk factors and thus the probability of several such factors coinciding, as in case of the Tesla accident, thereby decreasing the risk for an accident. To avoid focusing on unrealistically extreme scenarios while overlooking much more likely and mundane factors, the scenarios should be representative for reality \cite{pfefferSzenariobasierteSimulationsgestuetzteFunktionale2020, wachenfeldFreigabeAutonomenFahrens2015}. To achieve a high coverage of the space of all possible scenarios, and to avoid overfitting the function to certain selected conditions, a high number of different scenarios should be tested \cite{pfefferSzenariobasierteSimulationsgestuetzteFunktionale2020, wachenfeldFreigabeAutonomenFahrens2015}. A proposed approach, taken by the AVEAS project, is therefore to start from real data and extract scenarios from those. These are then parametrized and modified \cite{montanariManeuverbasedResimulationDriving2021, montanariPatternRecognitionDriving2020, neurohrFundamentalConsiderationsScenarioBased2020, ponnOPTIMIZATIONBASEDMETHODIDENTIFY2019, wachenfeldFreigabeAutonomenFahrens2015} to achieve higher variety. Current scenario descriptions are mainly trajectory-based \cite{karunakaranAutomaticLaneChange2022, parkCreatingDrivingScenarios2020, wagnerCrossVerificationUseSimulation2019, xinxinCSGCriticalScenario2020, zofkaDataDrivenSimulationParametrization2015}. An advancement of this is the \emph{maneuver-based} approach \cite{montanariManeuverbasedResimulationDriving2021, kolbAutomaticEvaluationAutomatically2022, heinzTrackScenariobasedTrajectory2017}, which forms the basis for this paper.

After the general background on scenario-based testing of AD functions in simulations given in this section, some wording will be discussed in Sec.~\ref{Wording}. Then, the results of the literature review will be presented in Sec.~\ref{LitReview}. The aim is to assess the current state of the art in the field of maneuver-based scenario description. Available work that can serve as foundation for further research in this field will be outlined and categorized, and current limitations of the presented methods, leaving room for further research, will be identified. 
Key conclusions are drawn in Sec.~\ref{Conclusion}.

\section{Wording}\label{Wording}
This work uses the classical definitions of \emph{scene} and \emph{scenario} as coined by Ulbrich et al. \cite{ulbrichDefiningSubstantiatingTerms2015}. Of special relevance for this paper is the term \emph{scenery}, referring to all geo-spatially stationary aspects of a scene such as the road network, signs, buildings and trees \cite{ulbrichDefiningSubstantiatingTerms2015}.

\emph{Resimulation} here refers to the simulation of scenarios that are derived from real-world driving data. Thus, it covers not only a replay of the recorded scenario but also the simulation of scenarios obtained by variation of the original scenario.

Within this paper and the ones referenced below, the term \emph{maneuver} is considered to be part of the system state, as opposed to part of the transition between quasi-stationary system states (see \cite{jatzkowskiFahrmanoeverbegriffImKontext2021} for more details). This is also the understanding dominating in literature.
According to Braun et al. \cite{braunCollectionRequirementsModelbased2021}, maneuvers can be categorized based on their direction, the presence of interaction with other traffic participants, their start and end conditions as well as their hierarchical level (see below). Start conditions can be either \emph{triggered} or \emph{implicit}, end conditions \emph{self-terminating} or \emph{indefinite}. Regarding the hierarchical level, a maneuver can be categorized as either \emph{composed} or \emph{atomic}. When putting the driver into the center of consideration, maneuvers can also be classified according to their reason \cite{songSurroundingVehiclesLane2022}: they can be \emph{intended} or \emph{unintended} (for example caused by distraction or fatigue). 
Furthermore, a maneuver can be classified according to its level of abstraction, also coming from a driver's perspective: a maneuver can be \emph{strategic} (e.g. turning), \emph{tactical} (e.g. overtaking) or \emph{operative} (e.g. braking) \cite{dongesASPEKTEAKTIVENSICHERHEIT1982}. Already this short consideration of the term maneuver hints at difficulties that arise when trying to set up a maneuver-based scenario description. 
Open questions that need to be clarified are:
\begin{itemize}
\item Which maneuvers -- in the sense of general kind, hierarchical level and level of abstraction -- to choose? 
\item Which start and end conditions to select?
\end{itemize}
For the sake of flexibility, an advantageous choice of maneuvers could be a set of atomic behaviors from which every higher-level maneuver can be derived as proposed in \cite{jatzkowskiFahrmanoeverbegriffImKontext2021, braunCollectionRequirementsModelbased2021} and \cite{assadiFlexibleStochasticMicroscopic2020}. This would partially answer the first question. However, the decisions on which level of abstraction to define the maneuvers and which categorizations to apply are yet to be made. Jatzkowski~et~al.~\cite{jatzkowskiFahrmanoeverbegriffImKontext2021}, for example, propose to describe the vehicle's state at any time using two overlapping maneuvers, one for the lateral and one for the longitudinal direction.

\section{Literature Review}\label{LitReview}
This section presents the scenario descriptions currently used in resimulation and outlines research conducted in the two fields ``maneuver detection, classification and prediction'' and ``trajectory prediction and modeling'' that are capable of providing methods for the maneuver-based approach.

The motivation behind this selection of research fields is the following: a key advantage of maneuver-based resimulation (see Sec.~\ref{Montanari}) is the possibility to easily vary the scenario's course or its scenery. These variations require conditional models enabling the derivation of consistent low-level details (e.g. trajectories) based on high-level maneuvers. Maneuver-based resimulation is a relatively new tool in the field of virtual validation for AD. Thus, few models exist that have been directly developed for this application. However, yet a wide range of models supports similar tasks, in particular the inverse type of task, such as predicting the upcoming trajectory or detecting a maneuver based on low-level details. Therefore, this literature review covers these fields as well.

All papers whose models are presented within this section are categorized using the scheme defined by Tab. \ref{tab:paper_overview}. The aim is to assess the potential of the presented methods to be used for maneuver-based scenario descriptions in the context of AD simulations. The categorizations applied within Tab. \ref{tab:paper_overview}  cover different aspects relevant for this assessment.
The row and column a method is assigned to describe the general model characteristics. The row indicates the granularity of the fitting data used for parametrization/calibration of the models and thus indicates down to which level the realistic behavior of the model is ensured. 
The columns indicate up to which granularity phenomena are considered within the model descriptions. Both aspects need to be in-line with the requirements on the models needed for a maneuver-based scenario description. 
On-board details are quantities related to the vehicle's interior that are not typically visible from the outside. For the granularity of the model descriptions they are further separated into vehicle dynamics-related, human factor-related, or both. Vehicle dynamics-related on-board details include pedal positions, brake pressure, as well as slip and steering dynamics of the vehicle; those related to human factors refer to the physical and psychological state of the driver comprising quantities such as the driver's head position, perception thresholds, level of aggressiveness or personality traits. 
The color scheme indicates whether or not a method differentiates between maneuvers, and if so, whether it is able to detect and classify these maneuvers. The presence of the maneuver concept indicates a familiarity to the maneuver-based approach and thus bears the potential for reuse.
Methods that are applied in the context of AD simulations are highlighted with bold letters. Methods of this category are thus already usable within AD simulations and possibly also when using the maneuver-based approach. 
All categorizations are made based on what was actually used in the publication, as opposed to what could in theory be possible (e.g. fitting the model with data of lower granularity than the ones used in the paper). 

This section continues with a review of all models listed in Tab. \ref{tab:paper_overview} and concludes with a synopsis in Sec. \ref{sec:Synopsis} that presents the insights that can be gained from Tab. \ref{tab:paper_overview} against the background of the conducted literature review.

\newcommand\myparbox[1]{\parbox{0.125\textwidth}{#1}}
\newcommand\myparboxlegend[1]{\parbox{16cm}{\vspace{3pt}#1}}
\newcommand\myparboxlegendx[1]{\parbox{16cm}{#1}}
\newcommand\mysmallerparbox[1]{\parbox{0.125\textwidth}{#1}}
\newcommand\myfirstcolumnparbox[1]{\parbox{0.08\textwidth}{\vspace{2pt}#1\vspace{2pt}}}
\newcommand\mysecondcolumnparbox[1]{\parbox{0.42\textwidth}{\vspace{2pt} #1 \vspace{2pt}}}
\newcommand\myspacingparbox[1]{\parbox{0.125\textwidth}{\vspace{2pt} #1 \vspace{2pt}}}
\newcommand\myspacingparboxx[1]{\parbox{0.125\textwidth}{\vspace{2pt} #1 \vspace{2pt}}}
\newcommand\mywideparboxx[1]{\parbox{6cm}{\vspace{2pt} #1}}
\newcommand\mywideparbox[1]{\parbox{4cm}{\vspace{3pt} #1 \vspace{2pt}}}
\newcommand\noman[1]{\textcolor[HTML]{fec44f}{\cite{#1}}}
\newcommand\manspec[1]{\textcolor[HTML]{ED7FC6}{\cite{#1}}}
\newcommand\man[1]{\textcolor[HTML]{5D529C}{\cite{#1}}}

\begin{table*}[htbp]
\caption{Paper categorization}
\begin{center}
\begin{tabular}{lllll}
\hline
\multicolumn{1}{l|} {\multirow{3}{*}{\textbf{\myfirstcolumnparbox{Granularity of fitting data down to...}}}}
& \multicolumn{4}{c}{\textbf{\mywideparboxx{Granularity of model descriptions down to…\vspace{2pt}}}}\\ \cline{2-5} 
\multicolumn{1}{l|}{} & 
\multicolumn{1}{l|}{\multirow{2}{*}{\textit{\mysecondcolumnparbox{\centering Map-referenced trajectories}}}} & 
\multicolumn{3}{c}{\textit{{\mywideparbox{\centering On-board details}}}} \\ \cline{3-5}
\multicolumn{1}{l|}{} & \multicolumn{1}{l|}{} & 
\multicolumn{1}{l|}{\textit{\myspacingparbox{\centering vehicle dynamics}}}     & \multicolumn{1}{l|}{\textit{\myspacingparbox{\centering vehicle dynamics \&\\ human factors}}}& 
\multicolumn{1}{l}{\textit{\myspacingparbox{\centering human factors}}}     \\
 \hline
\multicolumn{1}{l|}{\textit{\myfirstcolumnparbox{Traffic densities}}} &
  \multicolumn{1}{l|}{\mysecondcolumnparbox{\noman{rufGeometrieUndTopologie2018}}} &
  \multicolumn{1}{l|}{\myparbox{}} &
  \multicolumn{1}{l|}{\myparbox{
  \textbf{\manspec{hochstaedterSimulatingMacroscopicTraffic1998}}}} &
  \myspacingparboxx{\textbf{\manspec{krajzewiczTrafficSimulationSUMO2010}} \textbf{\manspec{fellendorfMicroscopicTrafficFlow2010}} \textbf{\manspec{sykesTrafficSimulationParamics2010}}\\ \textbf{\manspec{casasTrafficSimulationAimsun2010}} \textbf{\manspec{liuTrafficSimulationDRACULA2010}} \textbf{\manspec{yangNovelCarFollowingControl2019}}} \\
   \hline
\multicolumn{1}{l|}{\textit{\myfirstcolumnparbox{Map-referenced trajectories}}} &
  \multicolumn{1}{l|}{\mysecondcolumnparbox{\textbf{\man{montanariManeuverbasedResimulationDriving2021}}
\textbf{\man{karunakaranAutomaticLaneChange2022}}
\textbf{\noman{parkCreatingDrivingScenarios2020}}
\textbf{\noman{wagnerCrossVerificationUseSimulation2019}}
\textbf{\noman{xinxinCSGCriticalScenario2020}}
\textbf{\man{zofkaDataDrivenSimulationParametrization2015}}
\textbf{\man{kolbAutomaticEvaluationAutomatically2022}}
\textbf{\man{heinzTrackScenariobasedTrajectory2017}}
\man{gaoAdaptiveWindowSize2020}
\manspec{fengSupportVectorMachine2022}
\man{zhangHybridApproachTurning2020}
\man{zhangEnsembleLearningOnline2022}\\
\man{choiComparisonMachineLearning2021}
\man{benterkiMultiModelLearningBasedFramework2020}
\man{yuanLanechangePredictionMethod2018}
\manspec{deoConvolutionalSocialPooling2018}
\man{remmenCutinScenarioPrediction2018}
\noman{manttariLearningPredictLane2018}
\manspec{maResearchMultiobjectiveTrajectory2021}  
\man{deoHowWouldSurround2018}
\man{kasperErkennungFahrmanoevernMit2012}
\man{patelPredictingFutureLane2019}
\man{liDynamicBayesianNetwork2019}\\
\noman{benterkiLongTermPredictionVehicle2019}
\man{minRNNBasedPathPrediction2019}
\manspec{zhuCombinedHierarchicalLearning2021}
\noman{altcheLSTMNetworkHighway2017}
\noman{yiVehicleTrajectoryPrediction2015}
\manspec{butakovPersonalizedDriverVehicle2015}
\man{biparvaVideoActionRecognition2022}
\textbf{\manspec{assadiFlexibleStochasticMicroscopic2020}}
\manspec{petrichMapbasedLongTerm2013}
\textbf{\manspec{hubmannAutomatedDrivingUncertain2018}}
\textbf{\manspec{qiStochasticLateralNoise2022}}\hspace{0.55cm}}}&
  \multicolumn{1}{l|}{\myparbox{\textbf{\manspec{kumarMultilevelModelingTraffic2014}}}} &
  \multicolumn{1}{l|}{\myparbox{\textbf{\manspec{jamiAugmentedDriverBehavior2022}}}} &
  \myparbox{\textbf{\manspec{chenCalibrationMITSIMIDM2010}} \textbf{\manspec{kurtcCalibratingLocalPlatoon2016}} \manspec{xieDatadrivenLanechangingModel2019}} \\
   \hline
\multicolumn{1}{l|}{\textit{\myfirstcolumnparbox{On-board details}}} &
  \multicolumn{1}{l|}{\mysecondcolumnparbox{\manspec{zhuPersonalizedLaneChangeAssistance2018}}} &
  \multicolumn{1}{l|}{\myparbox{\textbf{\manspec{xuAutomatedLearningBasedProcedure2019}}}} &
  \multicolumn{1}{l|}{\myparbox{\manspec{leonhardtFeatureEvaluationLane2017}\textbf{\manspec{friesDriverBehaviorModel2022}}}} &
  \myparbox{\man{gaoSelfSupervisedLearningDriving2020} \man{huDecisionTreeBasedManeuver2017} \man{ouDeepLearningBasedDriving2020}} \\
\hline
\multicolumn{5}{l}{
\begin{tabular}[c]{@{}l@{}}
{\myparboxlegend{\textcolor[HTML]{fec44f}{Method does not differentiate between maneuvers} $\vert$ \textcolor[HTML]{ED7FC6}{Method differentiates between maneuvers but is not able to detect and classify them}}}\\
\myparboxlegendx{\textcolor[HTML]{5D529C}{Method differentiates between maneuvers and is able to detect and classify them} $\vert$ \textbf{Bold letters indicate methods applied in AD simulations}}\end{tabular}} 
\end{tabular}%
\label{tab:paper_overview}
\end{center}
\vspace{-0.6cm}
\end{table*}

\subsection{Resimulation}\label{Resimulation}
This section explains the scenario description approaches currently used in resimulations and presents their advantages as well as limitations.
\subsubsection{Trajectory-based approach}
The majority of literature published in the field of scenario-based simulation uses a trajectory-based scenario description. Hereby, a georeferenced trajectory is generated and precisely reconstructed in a simulation environment based on the measured coordinates and relative positions of the ego vehicle and the surrounding traffic participants \cite{karunakaranAutomaticLaneChange2022, parkCreatingDrivingScenarios2020, wagnerCrossVerificationUseSimulation2019, xinxinCSGCriticalScenario2020, zofkaDataDrivenSimulationParametrization2015}. On the one hand, this enables an easy generation of trajectory-based scenarios, only requiring the ability to track the GNSS coordinates (or comparable) of a traffic participant. The accuracy of the resulting trajectory is only limited by the measurement accuracy and can include detailed lateral offset as well as velocity profiles \cite{montanariManeuverbasedResimulationDriving2021}. Therefore, the trajectory-based approach is suitable in case the aim is to exactly resimulate the course of a recorded scenario. On the other hand, the utilization of coordinates entails the limitation that, by this, the scenario is bound to the road geometry of the scenery in which it was recorded \cite{heinzTrackScenariobasedTrajectory2017}. Only the remaining aspects of the scenery such as buildings, colors or the weather can be changed. Moreover, a scenario described via a trajectory is difficult to vary as the trajectory's internal consistency and the consistency with the scenery must be maintained \cite{montanariManeuverbasedResimulationDriving2021}. This makes it difficult to change a vehicle's driving behavior based on the surrounding traffic or scenery elements. Some of the approaches also introduce the term ``maneuver'', e.g. to define variations of the trajectory \cite{zofkaDataDrivenSimulationParametrization2015} or limit the scope of the extracted trajectory \cite{karunakaranAutomaticLaneChange2022}.

\subsubsection{Maneuver-based approach}\label{Montanari}
In case of the maneuver-based approach, a recorded trajectory is generalized into maneuvers. This allows the transfer of a scenario to another start position within a given scenery, and from one scenery to another. The road geometry can be varied. 
Exemplary transfers of interest are listed below:
\begin{itemize}
\item shift of a scenario from a rural to an urban environment, potentially characterized by other lighting effects such as dazzling lights and different classes of traffic participants, for example pedestrians
\item replay of a cut-in scenario within a curve instead of a straight section of a highway, as occurring global velocity distributions and occlusion effects might differ
\item use of a left-hand instead of right-hand traffic scenery to test if the system is still able to operate safely in countries with left-hand traffic
\end{itemize}
Besides the transferability, the maneuver-based approach allows for easier parametrization and modification \cite{montanariManeuverbasedResimulationDriving2021}. In this work, parametrization refers to a variation of a quantity describing a maneuver such as the longitudinal distance to the vehicle in front before initiating a lane change. A modification, instead, changes the general course of a scenario by adding or deleting maneuvers or changing their order. Thus, by using a maneuver-based approach changes are not limited to small local variations as in case of the trajectory-based approach. Instead, it allows for variations of the scenario's general course -- in accordance with the possibly changed scenery or surrounding traffic. Due to the previously outlined need for creating a variety of scenarios for scenario-based testing, the maneuver-based approach bears the potential of providing a better and more profound foundation for the validation and verification of AD functions. However, the generalization introduced by the maneuver-based approach also has limitations. Besides the two open questions identified earlier, the description as such is non-trivial. It requires semantics capable of describing any scenario observable in reality. First approaches in this direction are presented in \cite{bachModelBasedScenario2016} and given by the ASAM OpenSCENARIO\footnote{https://www.asam.net/standards/detail/openscenario/} standard. 
Moreover, there is the trade-off between generalization on the one hand and the ability to accurately describe occurring \emph{submicroscopic} phenomena such as velocity distributions and the vehicle's lateral offset within its lane on the other hand. A more detailed explanation of the level identifier ``submicroscopic'' will be given in Sec.~\ref{microscopic}.

Possibilities to apply the maneuver-based approach are presented in  \cite{montanariManeuverbasedResimulationDriving2021} and \cite{heinzTrackScenariobasedTrajectory2017}. The approach of \cite{montanariManeuverbasedResimulationDriving2021} is taken up in \cite{kolbAutomaticEvaluationAutomatically2022} and transferred from highway to intersection scenarios.
In all cases, the recorded trajectory is abstracted into a sequence of maneuvers without gaps. In \cite{montanariManeuverbasedResimulationDriving2021} and \cite{kolbAutomaticEvaluationAutomatically2022} the approach proposed in  \cite{jatzkowskiFahrmanoeverbegriffImKontext2021} is applied, and a vehicle's state is at any time defined by two overlapping maneuvers: one for the lateral and one for the longitudinal direction. 
In case of \cite{heinzTrackScenariobasedTrajectory2017}, instead, exactly one maneuver is assigned to a vehicle at any time. In \cite{heinzTrackScenariobasedTrajectory2017} and \cite{montanariManeuverbasedResimulationDriving2021} distances and absolute positions are used as maneuver trigger conditions. In addition to that, timestamps are used in \cite{montanariManeuverbasedResimulationDriving2021}. While positions and distances could reflect the level at which also a driver takes decisions and starts maneuvers, this does not apply for pure timestamps. When using absolute positions as trigger conditions, the conditions need to be modified as well when transferring the scenario from one scenery to another or within a scenery.
The methods presented in \cite{montanariManeuverbasedResimulationDriving2021} and \cite{heinzTrackScenariobasedTrajectory2017} both simplify the submicroscopic behavior (see Sec.~\ref{microscopic}). Nevertheless, even the transfer of a simplified lateral offset profile of a scenario from one scenery to another yielded non-realistic behavior, as the lateral offset behavior depends on the curvature of the road and is therefore not transferable one-to-one \cite{heinzTrackScenariobasedTrajectory2017}. 
This addresses the general open question of how to perform a realistic parametrization or modification of a scenario. As long as a pure resimulation is performed, as done in \cite{montanariManeuverbasedResimulationDriving2021}, no issue arises. However, if the scenario is transferred to a scenery where for example the distance to be travelled during a maneuver is longer, the maneuver has to reflect this. Possibilities to achieve this are a stretching or a repetition of the maneuver or the increase of the velocity. However, the representativity of these approaches with respect to real-world driving behavior is thus far not substantiated by a systematic evaluation on real-world data.

\subsubsection{Microscopic and submicroscopic traffic simulation}\label{microscopic}
An alternative approach to describe the behavior of vehicles within simulations is the use of models as used in microscopic traffic simulations such as SUMO \cite{krajzewiczTrafficSimulationSUMO2010}, Vissim \cite{fellendorfMicroscopicTrafficFlow2010}, Aimsun \cite{casasTrafficSimulationAimsun2010}, Paramics \cite{sykesTrafficSimulationParamics2010} and DRACULA \cite{liuTrafficSimulationDRACULA2010}. Even though not commonly considered a maneuver-based approach, they can provide applicable models. The route a vehicle takes through the road network is determined on a high level through its origin and destination \cite{krajzewiczTrafficSimulationSUMO2010, fellendorfMicroscopicTrafficFlow2010, sykesTrafficSimulationParamics2010}.
 The description of the vehicle's microscopic behavior is commonly split in a lane-changing model for the lateral movement between lanes, and a car-following model for the longitudinal direction \cite{casasTrafficSimulationAimsun2010, krajzewiczTrafficSimulationSUMO2010, fellendorfMicroscopicTrafficFlow2010,liuTrafficSimulationDRACULA2010} such as the Wiedemann model in Vissim \cite{fellendorfMicroscopicTrafficFlow2010} or the Krauß model in SUMO \cite{krajzewiczTrafficSimulationSUMO2010}. In longitudinal direction, a vehicle chooses its velocity depending on the velocity of the leading vehicle, the distance to it and further ego vehicle-specific quantities such as its maximum deceleration ability \cite{krajzewiczTrafficSimulationSUMO2010, fellendorfMicroscopicTrafficFlow2010} as well as infrastructure-related criteria \cite{krajzewiczTrafficSimulationSUMO2010, casasTrafficSimulationAimsun2010} and the surrounding traffic \cite{casasTrafficSimulationAimsun2010}. To model the imperfection of a human driver, different strategies are applied. In SUMO \cite{krajzewiczTrafficSimulationSUMO2010} a stochastic deceleration is introduced, in Vissim \cite{fellendorfMicroscopicTrafficFlow2010} the following vehicle does not accelerate and decelerate directly but after predefined perception thresholds are reached. However, in all cases collisions between vehicles are excluded by design. 
The lane change behavior of a vehicle is modeled based on the tactical and strategic decisions of a driver: a vehicle changes the lane if this is required to reach its destination on the prescribed route or if it has an advantage by changing the lane (e.g. proceeding faster)  \cite{krajzewiczTrafficSimulationSUMO2010, fellendorfMicroscopicTrafficFlow2010, sykesTrafficSimulationParamics2010, casasTrafficSimulationAimsun2010, liuTrafficSimulationDRACULA2010}. For the modeling of the lane change itself, a simplified trajectory is used \cite{fellendorfMicroscopicTrafficFlow2010, sykesTrafficSimulationParamics2010}. 
Alternatives to the approaches presented so far apply machine learning methods to learn the models describing the longitudinal and lateral movement of a vehicle \cite{yangNovelCarFollowingControl2019, assadiFlexibleStochasticMicroscopic2020}.
Apart from the lane changes, the majority of the presented methods neglects lateral movements with respect to the road's center line \cite{casasTrafficSimulationAimsun2010, krajzewiczTrafficSimulationSUMO2010, liuTrafficSimulationDRACULA2010, sykesTrafficSimulationParamics2010}. An exception is Vissim \cite{fellendorfMicroscopicTrafficFlow2010} where the lateral position of a vehicle within its lane is based on the maximum longitudinal time-to-collision to the vehicle in front and a lateral safety distance to vehicles on the neighboring lanes. Under free driving conditions, it can be preset to rather left, right, center or random. Moreover, there are papers that explicitly deal with modelling the lateral displacement of a vehicle within its lane \cite{qiStochasticLateralNoise2022}.
To improve the realism on this level, so called \emph{submicroscopic} or \emph{nanoscopic} simulators are developed \cite{hoogendoornStateoftheartVehicularTraffic2001, ni2DSIMPrototypeNanoscopic2003, kumarMultilevelModelingTraffic2014, hochstaedterSimulatingMacroscopicTraffic1998, jamiAugmentedDriverBehavior2022}. The work in this field can be split into two parts: the first deals with physics-based models for increased realism in vehicle dynamics \cite{mullakkal-babuHybridSubmicroscopicMicroscopicTraffic2021, ni2DSIMPrototypeNanoscopic2003, kumarMultilevelModelingTraffic2014, hochstaedterSimulatingMacroscopicTraffic1998, jamiAugmentedDriverBehavior2022}, while the second aims for setting up realistic behavior models for the driver by taking into account psychological driver characteristics such as the level of aggressiveness \cite{ni2DSIMPrototypeNanoscopic2003, hochstaedterSimulatingMacroscopicTraffic1998, jamiAugmentedDriverBehavior2022}. In all cases the submicroscopic models are coupled with microscopic ones. While the latter ones determine the long-term behavior of the vehicles, the former ones define the details of execution. 

In the field of microscopic traffic simulation, the driver's behavior and the decisions behind it are described on a high level. However, one problem that the approaches referenced before have in common is that they are calibrated to get a representative macroscopic traffic flow \cite{krajzewiczTrafficSimulationSUMO2010,casasTrafficSimulationAimsun2010,liuTrafficSimulationDRACULA2010}, while the individual vehicle behavior might not be realistically modeled. To solve this issue, some recent approaches use trajectory data for calibration \cite{chenCalibrationMITSIMIDM2010,kurtcCalibratingLocalPlatoon2016}. 
Another aspect revealing a lack of realism is the complete absence of collisions enforced through the model design. 
Consequently, these approaches are not sufficient when aiming for the immediate safety evaluation of an AD function such as a cut-in detection. When using a simulator with vehicles always driving in the center of the lane for the training of a cut-in detection, the algorithm would learn that whenever a vehicle is slightly leaving the center of the lane, it will definitely change the lane. Consequently, using such a model for validation would fail to evaluate the actual capabilities of a given detector, leaving substantial uncertainty to what degree the function is safe under realistic traffic conditions. 
Similarly, it is not possible to draw a conclusion on whether a system could have avoided an accident if these were in beforehand excluded by the model.

\subsection{Maneuver detection, prediction and classification}\label{manpred}
The aim of research performed in this field is the detection of maneuvers in data such as real-world vehicle trajectories and their classification. Both tasks can be either performed online while the maneuver takes place, or offline and retrospectively on the full driving data. If the goal is to detect a maneuver before it starts or immediately after it has started, it is also referred to as maneuver prediction \cite{gaoAdaptiveWindowSize2020, gaoSelfSupervisedLearningDriving2020, huDecisionTreeBasedManeuver2017, biparvaVideoActionRecognition2022}. Insights gained and methods developed in the field of maneuver detection and classification could provide a fruitful ground for the advancement of the maneuver-based approach. Therefore, the state of the art is assessed within this section. Methods for maneuver detection and classification are developed since driver assistance systems need to have high situational awareness of the ego vehicle's environment as well as the driver's state \cite{gaoAdaptiveWindowSize2020}. For this, it is important for the system to predict maneuvers of the ego vehicle as well as surrounding vehicles. By this, an AD function is enabled to align its reactions with the intended behavior of the driver \cite{fengSupportVectorMachine2022, ouDeepLearningBasedDriving2020}, to remind the driver of actions that need to be taken \cite{gaoSelfSupervisedLearningDriving2020}, to warn him or her in case the intended maneuver cannot be executed safely \cite{gaoAdaptiveWindowSize2020} or just keep the driver informed about the environment \cite{ranaPartiallyVisibleLane2022}. Examples for maneuvers whose detection and classification is desired apart from the road following maneuver are turning maneuvers \cite{gaoSelfSupervisedLearningDriving2020, zhangHybridApproachTurning2020, zhangEnsembleLearningOnline2022, gaoAdaptiveWindowSize2020, ouDeepLearningBasedDriving2020}, lane change maneuvers \cite{biparvaVideoActionRecognition2022, ouDeepLearningBasedDriving2020, gaoSelfSupervisedLearningDriving2020, fengSupportVectorMachine2022, gaoConditionalArtificialPotential2019, choiComparisonMachineLearning2021, benterkiMultiModelLearningBasedFramework2020, zhangEnsembleLearningOnline2022, yuanLanechangePredictionMethod2018, deoConvolutionalSocialPooling2018, remmenCutinScenarioPrediction2018, manttariLearningPredictLane2018, gaoAdaptiveWindowSize2020, leonhardtFeatureEvaluationLane2017, kasperErkennungFahrmanoevernMit2012, maResearchMultiobjectiveTrajectory2021}, 
braking \cite{deoConvolutionalSocialPooling2018}, and drifts into another lane \cite{deoHowWouldSurround2018}. 
For the detection and classification of these maneuvers, a wide variety of methods such as clustering algorithms \cite{zhangEnsembleLearningOnline2022, maResearchMultiobjectiveTrajectory2021, zhuPersonalizedLaneChangeAssistance2018}, support vector machines \cite{fengSupportVectorMachine2022, choiComparisonMachineLearning2021} hidden Markov models \cite{yuanLanechangePredictionMethod2018, deoHowWouldSurround2018}, neural networks \cite{manttariLearningPredictLane2018,gaoAdaptiveWindowSize2020, ouDeepLearningBasedDriving2020, choiComparisonMachineLearning2021, benterkiMultiModelLearningBasedFramework2020, ouDeepLearningBasedDriving2020, biparvaVideoActionRecognition2022, patelPredictingFutureLane2019}, dynamic Bayesian networks \cite{liDynamicBayesianNetwork2019, kasperErkennungFahrmanoevernMit2012}, decision trees \cite{huDecisionTreeBasedManeuver2017} and random forest \cite{choiComparisonMachineLearning2021} are applied. 
Just as the used algorithms, also the data fed to those in order to detect and classify maneuvers are diverse.
The predominantly used features are physical quantities, especially the longitudinal and lateral coordinates of the vehicle for which the maneuver is expected \cite{zhangHybridApproachTurning2020, maTrafficPredictTrajectoryPrediction2019, deoConvolutionalSocialPooling2018, remmenCutinScenarioPrediction2018, fengSupportVectorMachine2022, maResearchMultiobjectiveTrajectory2021, gaoConditionalArtificialPotential2019, kasperErkennungFahrmanoevernMit2012}, referred to as target vehicle. 
Beside the position information, also the longitudinal and lateral velocity and acceleration of the target vehicle are of importance \cite{zhangHybridApproachTurning2020, yuanLanechangePredictionMethod2018, zhangEnsembleLearningOnline2022, remmenCutinScenarioPrediction2018, gaoConditionalArtificialPotential2019, benterkiMultiModelLearningBasedFramework2020, leonhardtFeatureEvaluationLane2017, kasperErkennungFahrmanoevernMit2012}. 
Further physical and vehicle-related quantities used for the detection and classification of maneuvers are the yaw angle \cite{zhangHybridApproachTurning2020, benterkiMultiModelLearningBasedFramework2020, kasperErkennungFahrmanoevernMit2012}, the yaw rate \cite{benterkiMultiModelLearningBasedFramework2020, liDynamicBayesianNetwork2019, leonhardtFeatureEvaluationLane2017, kasperErkennungFahrmanoevernMit2012}, turn and brake indicators \cite{liDynamicBayesianNetwork2019, leonhardtFeatureEvaluationLane2017}, pedal positions \cite{leonhardtFeatureEvaluationLane2017} and dimensions \cite{kasperErkennungFahrmanoevernMit2012} of the target vehicle. 
Features that can be fed into the model to also include the road geometry and topology are the curvature of the road and the existence of a left or right lane next to the target vehicle \cite{liDynamicBayesianNetwork2019, ouDeepLearningBasedDriving2020, kasperErkennungFahrmanoevernMit2012}. To take into account the surrounding traffic, relative positions, velocities and accelerations between vehicles \cite{maResearchMultiobjectiveTrajectory2021, zhangEnsembleLearningOnline2022, liDynamicBayesianNetwork2019, leonhardtFeatureEvaluationLane2017} are considered. A step further is made in \cite{deoConvolutionalSocialPooling2018}, \cite{patelPredictingFutureLane2019} and \cite{remmenCutinScenarioPrediction2018} where also the overall traffic situation is considered for the maneuver detection and classification. In \cite{biparvaVideoActionRecognition2022} the surrounding traffic as well as road conditions are included implicitly by using video data. When only focusing on the detection and classification of ego vehicle maneuvers, further features of the driver such as physiological signals \cite{gaoSelfSupervisedLearningDriving2020}, the head movement \cite{ouDeepLearningBasedDriving2020, leonhardtFeatureEvaluationLane2017} and personality traits \cite{huDecisionTreeBasedManeuver2017} can be used. 

As outlined, a great variety of maneuver sets, methods for their detection and classification and features fed to these methods exists in literature. The used maneuvers are characterized by specific patterns that can be found in the features. This is exploited by the applied methods. To achieve a realistic modeling of the vehicle's submicroscopic behavior in a simulation, these characteristic patterns need to be reproduced under the given circumstances such as the ego vehicle's state, the surrounding traffic and the road structure.
The maneuvers used in the cited papers can give orientation for the selection of a maneuver set for the scenario description and the distinction between maneuvers. Thus, they can help to answer the open questions identified in Sec.~\ref{Wording}. For instance, it needs to be clarified when a lane change starts and ends. While there is a common understanding that the ``middle'' of the lane change is the crossing of the lane marking \cite{biparvaVideoActionRecognition2022, benterkiMultiModelLearningBasedFramework2020}, there are different understandings concerning its start and end. In some cases, a lane change is defined by a time window around the crossing of the lane marking \cite{benterkiMultiModelLearningBasedFramework2020}, in others by a spatial area \cite{assadiFlexibleStochasticMicroscopic2020} or the start and end of lateral movement~\cite{butakovPersonalizedDriverVehicle2015}.

\subsection{Trajectory prediction and modeling}\label{sec:traj_pred}
Trajectory prediction is closely related to maneuver detection and classification. In contrast to maneuver prediction, trajectory prediction does not aim for predicting the next maneuver of a vehicle out of a discrete maneuver set, but its detailed upcoming course \cite{schlechtriemenWhenWillIt2015}. This, in turn, usually depends on the maneuver: while a lane keeping maneuver yields a trajectory staying on the vehicle's current lane, an upcoming lane change needs to be followed by a trajectory continuing on a neighboring lane. Therefore, trajectory prediction is often preceded by a maneuver prediction which determines the selection of the maneuver-dependant trajectory prediction algorithm \cite{maResearchMultiobjectiveTrajectory2021}. Also these techniques find application in AD functions. They are for example applied to target vehicles to plan the ego vehicle's trajectory accordingly \cite{maResearchMultiobjectiveTrajectory2021, benterkiLongTermPredictionVehicle2019, benterkiMultiModelLearningBasedFramework2020, maTrafficPredictTrajectoryPrediction2019, deoHowWouldSurround2018, minRNNBasedPathPrediction2019, rufGeometrieUndTopologie2018} or to the ego vehicle, e.g. to adapt the driving style of an AD function to the one of the driver to increase the driver's comfort \cite{zhuCombinedHierarchicalLearning2021, butakovPersonalizedDriverVehicle2015}. Besides the field of trajectory \emph{prediction} for real-world situations, there is also the field of trajectory \emph{modeling}, aiming for generating a realistic vehicle trajectory within a simulation \cite{friesDriverBehaviorModel2022, xieDatadrivenLanechangingModel2019, xuAutomatedLearningBasedProcedure2019}.
A frequently applied method for trajectory prediction and modeling are long short-term memory networks \cite{maResearchMultiobjectiveTrajectory2021, xieDatadrivenLanechangingModel2019, maTrafficPredictTrajectoryPrediction2019, altcheLSTMNetworkHighway2017, xuAutomatedLearningBasedProcedure2019}, however, also other variants of neural networks \cite{benterkiMultiModelLearningBasedFramework2020, zhuCombinedHierarchicalLearning2021, benterkiLongTermPredictionVehicle2019, minRNNBasedPathPrediction2019, xuAutomatedLearningBasedProcedure2019} as well as Gaussian Mixture Models \cite{deoHowWouldSurround2018, butakovPersonalizedDriverVehicle2015}, polynomial fitting \cite{yiVehicleTrajectoryPrediction2015} and stochastic methods \cite{friesDriverBehaviorModel2022, rufGeometrieUndTopologie2018, petrichMapbasedLongTerm2013, hubmannAutomatedDrivingUncertain2018} are used. Depending on the application, in some cases the goal is to construct an exact trajectory \cite{maTrafficPredictTrajectoryPrediction2019, altcheLSTMNetworkHighway2017, yiVehicleTrajectoryPrediction2015, friesDriverBehaviorModel2022, friesDriverBehaviorModel2022, xieDatadrivenLanechangingModel2019, xuAutomatedLearningBasedProcedure2019}, for example when aiming for a personalized driving function or a trajectory within a simulation, in other cases the aim is to return a set of possible trajectories, combined with a probability for each of them, or a fused trajectory \cite{maResearchMultiobjectiveTrajectory2021, deoHowWouldSurround2018, rufGeometrieUndTopologie2018, petrichMapbasedLongTerm2013, hubmannAutomatedDrivingUncertain2018}. This becomes important when the aim is to predict the trajectory of another vehicle to plan the ego vehicle's behavior. 
The prediction of the upcoming trajectory is based on the previous trajectory \cite{benterkiMultiModelLearningBasedFramework2020, maResearchMultiobjectiveTrajectory2021, maTrafficPredictTrajectoryPrediction2019, minRNNBasedPathPrediction2019, altcheLSTMNetworkHighway2017, petrichMapbasedLongTerm2013} as well as velocity and acceleration data \cite{zhuCombinedHierarchicalLearning2021, benterkiLongTermPredictionVehicle2019, minRNNBasedPathPrediction2019, altcheLSTMNetworkHighway2017, yiVehicleTrajectoryPrediction2015, butakovPersonalizedDriverVehicle2015, petrichMapbasedLongTerm2013}. As in the case of maneuver prediction, also in this case surrounding traffic \cite{benterkiLongTermPredictionVehicle2019, altcheLSTMNetworkHighway2017, butakovPersonalizedDriverVehicle2015, hubmannAutomatedDrivingUncertain2018} and the road structure \cite{minRNNBasedPathPrediction2019, yiVehicleTrajectoryPrediction2015, petrichMapbasedLongTerm2013} are taken into account. There are also approaches in which the modeling goes down to driver \cite{friesDriverBehaviorModel2022} and vehicle level \cite{friesDriverBehaviorModel2022, xuAutomatedLearningBasedProcedure2019} meaning that for example the driver's perception and pedal position are taken into account. 
The outcome of the trajectory prediction is manifold. Apart from a direct prediction of the next position(s) \cite{benterkiMultiModelLearningBasedFramework2020, altcheLSTMNetworkHighway2017, yiVehicleTrajectoryPrediction2015}, there exist methods predicting the lateral and longitudinal trajectory separately \cite{benterkiLongTermPredictionVehicle2019} or delivering a chain of velocities or accelerations from which the positions can be derived \cite{deoHowWouldSurround2018, altcheLSTMNetworkHighway2017, hubmannAutomatedDrivingUncertain2018}. Due to the usual dependence of the trajectory on the upcoming maneuver, there are also methods dealing with trajectory prediction for a certain maneuver such as a lane change \cite{zhuCombinedHierarchicalLearning2021, minRNNBasedPathPrediction2019, butakovPersonalizedDriverVehicle2015}.

As in case of trajectory prediction, also when deriving models for the submicroscopic vehicle behavior, the aim is to describe the vehicle's dynamic quantities with a high temporal and spatial resolution. However, the focus of both methods differs. In case of trajectory prediction, the aim is to get a high accuracy for the immediately upcoming vehicle behavior while the goal of a submicroscopic behavior model is to get a realistic behavior of the vehicle for the full duration of a maneuver. Furthermore, prediction tasks typically aim to optimize extrinsic quantities, such as providing a trajectory of minimum expected error w.r.t. all possible outcomes, or a conservative estimate maximizing safety for the AD function, whereas, when used for simulation, submicroscopic behavior models aim at sampling realistic candidates from the modeled distribution, including candidates with low likelihood. Measuring the degree of realism of modeled behavior, and determining the required degree of realism for a given simulation task, is an open question in its own. As the methods developed within the field of trajectory prediction are limited to short-term prediction, they are not suitable for predicting the course of a vehicle for the full duration of a road following maneuver. However, they can be used to vary the course a vehicle takes during a maneuver that -- in case of a simulation -- is known in advance, to get a realistic imperfect submicroscopic driving behavior. Due to the comparatively low duration of a lane change maneuver, these maneuvers could be an exception and trajectory prediction models might be applicable. The methods for trajectory modeling directly address the task of providing a representative trajectory and therefore are promising candidates to build on further work in the field of submicroscopic behavior models.

\subsection{Synopsis}\label{sec:Synopsis}
All methods presented within this section have been placed in Tab.~\ref{sec:Synopsis}. Almost every combination of model granularity and fitting data granularity is covered. The predominant share of papers differentiates between maneuvers. Within those, the number of methods being able perform the detection and classification themselves and those not being able to do so is almost equal. As soon as the model granularity goes down to on-board details, all papers apply a maneuver differentiation. This could indicate that the relevance of maneuvers increases if on-board details are considered within the model. More than half of the reviewed papers' models reach down to a granularity of map-referenced trajectories and were fitted with data of that same granularity. This is exactly the minimum model and fitting data granularity needed for advancing the maneuver-based approach, as map-referenced trajectories represent the level visible to an AD function. To ensure realistic behavior of the model at this granularity, the fitting data should have at least the same granularity. Considering also human factors and vehicle dynamics-related phenomena in models for the maneuver-based approach and enriching the fitting data with on-board details could be useful to relate the characteristics observed in resulting trajectories to their in-vehicle origin and increase their representativity for real-world driving behavior. 

Only three of the papers listed in Tab.~\ref{tab:paper_overview} use the maneuver-based approach for resimulation -- with the limitations identified earlier. This hints at a research gap in the field of maneuver-based resimulation. However, as outlined earlier and demonstrated by the classification in Tab.~\ref{tab:paper_overview}, alternative resimulation approaches and methods for maneuver detection and classification and trajectory prediction and modeling address similar issues and thus provide a valuable foundation for an advancement of the maneuver-based approach.

\section{Conclusion}\label{Conclusion}
This paper presents the approach of using a maneuver-based scenario description within AD simulations and highlights its advantages as well as current limitations. The maneuver-based approach offers more degrees of freedom when varying a recorded scenario but its description is non-trivial. Prerequisites for a maneuver-based scenario description are the selection of a maneuver set and respective trigger conditions. To exploit the full potential of the maneuver-based approach the performed literature review indicates the need to use high-level maneuvers at the level on which a driver plans maneuvers, combined with submicroscopic behavior models and trigger conditions derived from the static environment or surrounding traffic participants. Based on this research gap, related research fields that can serve as starting point for further research were identified and categorized. 

\bibliography{bibliography} 

\end{document}